\documentclass[3p, twocolumn]{elsarticle}
\usepackage[utf8]{inputenc}

\usepackage{amsmath, amssymb, bm}
\usepackage{booktabs}
\usepackage{multirow}

\usepackage{tikz}
\usetikzlibrary{arrows}

\usepackage{xspace}
\usepackage{xcolor}
\usepackage{lineno}

\usepackage{array}
\newcolumntype{m}[1]{>{\centering\arraybackslash}p{#1}}

\usepackage{multirow}
\usepackage{rotating}
\usepackage{tikz}

\usepackage{amsmath,amssymb,amsfonts}
\usepackage{algorithmic}
\usepackage{graphicx}
\usepackage{textcomp}
\usepackage{xcolor}
\usepackage{subcaption}

\usepackage[colorlinks, allcolors=black]{hyperref}
\usepackage{cleveref}

\usepackage{lineno,hyperref}

\journal{Journal of \LaTeX\ Templates}

\begin{document}

%
%
%
%

%
%
%
%
%
%
%
%
%
%
%
%
%
%
%
%
%
%
%

\begin{frontmatter}

\title{Extending the Kinematic Theory of Rapid Movements with new Primitives}

\author[ulpgc]{Miguel A. Ferrer}
\author[ulpgc]{Moises Diaz\corref{mycorrespondingauthor}}
\cortext[mycorrespondingauthor]{Corresponding author}
\ead{moises.diaz@ulpgc.es}
\author[ulpgc]{Jose Juan~Quintana}
\author[ulpgc]{Cristina Carmona-Duarte}
\author[epm]{Réjean Plamondon}

\address[ulpgc]{Instituto Universitario para el Desarrollo Tecnológico y la Innovación en Comunicaciones, Universidad de Las Palmas de Gran Canaria, Spain. Emails: \{miguelangel.ferrer, moises.diaz, josejuan.quintana, cristina.carmona \}@ulpgc.es}
\address[epm]{Polytechnique Montréal, Montréal, P.Q., Canada.  Email: rejean.plamondon@polymtl.ca.}

\begin{abstract}
The Kinematic Theory of rapid movements, and its associated Sigma-Lognormal, model 2D spatiotemporal trajectories. It is constructed mainly as a temporal overlap of curves between virtual target points.
 Specifically, it uses an arc and a lognormal as primitives for the representation of the trajectory and velocity, respectively. This paper proposes developing this model, in what we call the Kinematic Theory Transform, which establishes a mathematical framework that allows further primitives to be used. Mainly, we evaluate Euler curves to link virtual target points and Gaussian, Beta, Gamma, Double-bounded lognormal, and Generalized Extreme Value functions to model the bell-shaped velocity profile. Using these primitives, we report reconstruction results with spatiotemporal trajectories executed by human beings, animals, and anthropomorphic robots. 
\end{abstract}


\end{frontmatter}

 
\section{Introduction} 

 There exist many theories that have tried to describe the velocity profile of the movements of human beings in general and handwriting in particular\textcolor{black}{~\cite{de2019graphonomics}}. Specifically, \cite{plamondon2014recent} mentions models relying on neural networks, equilibrium point models, behavioral models, coupled oscillator models, differential equation models, kinematic models, and models exploiting minimization principles such as the minimization of the acceleration, or the energy, or the time, or the jerk, or the snap, or the torque changes or the sensory-motor noise. Many models exploit the properties of various mathematical functions to reproduce human movements: exponentials, second-order systems, Gaussians, beta functions, splines and trigonometrical functions.


Among the models which provide analytical representations, the kinematic theory of rapid human movements~\cite{plamondon1995kinematica} and its associated Sigma-Lognormal model have been extensively used to explain most of the basic phenomena reported in classical studies on human motor control and to study several factors involved in fine motricity\textcolor{black}{~\cite{pan2019characteristics, dentamaro2020gait,dentamaro2021fall}.}

 

 To work out the Sigma-Lognormal parameters, a spatiotemporal trajectory is transformed as a sequence of circumference arcs between virtual target points. A starting and an ending angle define each arc between virtual target points. Each ending virtual target point is the starting virtual target point of the next arc. Furthermore, each arc has a starting and ending time, but the finishing time of an arc is not the same as the starting time of the next arc. As a consequence, the arcs are temporally overlapped. Each arc is executed following a lognormal-shaped velocity curve and all the samples corresponding to a given time are vectorially summed to reconstruct the trajectory.

 As a result, the spatiotemporal trajectory can be analytically encoded into a sequence of virtual target points along with their starting and ending angles and their velocity parameters. As velocity and trajectory primitives, this process has traditionally used lognormal functions and arcs of circumference between virtual target points, respectively. From a purely mathematical point of view, such primitives can be changed, that is to say, the lognormal functions or the arcs can be substituted by other bell-shaped functions or other curves respectively. Insofar as a spatiotemporal trajectory is represented as a linear combination of weighted and shifted curves and bell-shaped functions, we propose as a further development a mathematical transform which also follows the paradigm of the Kinematic Theory of rapid movements.

 Hence, this paper defines what we call the Kinematic Theory Transform (KTT) for spatiotemporal trajectories with a bell-shaped velocity profile, which is often found in biological applications. More specifically, a mathematical framework is proposed to calculate their parameters and study the reach and range of KTT by processing signals produced by human, animal and even robot arms.

 With this purpose in mind, the Sigma-Lognormal model is completely reformulated to incorporate these extensions. In this way, Euler curves are added to improve the trajectory between virtual target points and additional bell-shaped functions are included to enhance the fitting of the bell-shaped velocity profile such as Gaussian, Beta, Gamma, Double-bounded lognormals and Generalized Extreme Value functions.

 The paper is organized as follows: 
 Section 2 establishes the mathematical framework to modify the arc between virtual target points by other analytical trajectories, while Section 3 introduces several bell-shaped functions that can be used instead of lognormals. The evaluation and related discussions are described in Section 4, while Section 5 concludes the article.

 \section{Kinematic Theory Transform}
 
 The Kinematic Theory Transform\footnote{For a more detailed introduction to this section, we refer readers to the brief review the Sigma-Lognormal model and iDeLog method, provided in the Supplemental file.} (KTT) is proposed as a useful tool for analyzing a wide range of spatiotemporal sequences, thus extending and extrapolating the applications of the Sigma-Lognormal model. The KTT is integrated into the iDeLog method~\cite{ferrer2018idelog}. This section presents the mathematical framework which enables the inclusion and use of new primitives in the KTT.

 \subsection{Generalizing the trajectory between virtual target points}
 
\begin{figure}
\centering
 \includegraphics[width=0.7\linewidth]{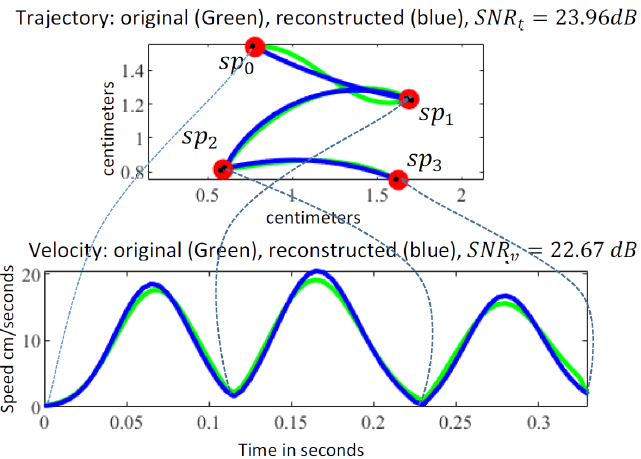}
  \caption{Example of a spatiotemporal sequence with an inflection point between $sp_0$ and $sp_1$ and just one stroke reconstructed with arcs.}
  \label{fig3} 
\end{figure}

 The arcs of circumference that link virtual points exhibit several limitations in the reconstruction of complex spatiotemporal trajectories. Several cases cannot be accurately reproduced by linking virtual target points with an arc of a circumference, for instance strokes that include inflexion points. This case is illustrated in Fig.~\ref{fig3}, in which the trajectory between $sp_0$ and $sp_1$ includes an inflexion point which happens to correspond to the maximum of the stroke in the bell-shaped velocity curve. The similarity between such a trajectory and an arc is poor, as can be seen in Fig.~\ref{fig3}.

The solution proposed by the KTT is to allow different ballistic trajectories between target points that include inflexion points, but without changing the stroke parameters which are the virtual target points $tp_{j-1}$ and $tp_j$ and the tangent angles $\theta_{sj}$ and $\theta_{ej}$. In this way we keep the biological meaning of the KTT.

To include a generalized trajectory in the KTT, let us define the trajectory by a general parametric curve: 

\begin{equation}
 \begin{split}
x_p & = f_j(u) \\
 y_p & = g_j(u)
\end{split}
\end{equation}

\noindent so that $tp_{j-1} = [f_j(u_{j1}), g_j(u_{j1})]$, $tp_{j} = [f_j(u_{j2}), g_j(u_{j2})]$, its derivative in $u=u_{j1}$ is $\tan(\theta_{sj})$ and its derivate in $u=u_{j2}$, $\tan(\theta_{ej})$. The length of the curve between the virtual targets points can be obtained as follows:

\begin{equation}\label{eq14}
 D_j = \displaystyle \int_{u_{1j}}^{u_{2j}} \sqrt{\left( \frac{\partial f_j(u)}{\partial u} \right)^2 + \left( \frac{\partial g_j(u)}{\partial u} \right)^2} \,du
\end{equation}


Then, the trajectory can be reconstructed as:


\begin{equation}\label{eq15}
 \vec{s}_r(t) = \displaystyle \sum_{j=1}^N \left[
f_j(u_j(t)); g_j(u_j(t)) \right]
\end{equation}

\noindent where $u_j(t)$ is the value that solves the equation:

\begin{multline}\label{eq16}
 \displaystyle \int_{u_{1j}}^{u_j(t)} \sqrt{\left( \frac{\partial f_j(u)}{\partial u} \right)^2 + \left( \frac{\partial g_j(u)}{\partial u} \right)^2} \,du =\\  D_j\displaystyle\int_0^t v_j(t; t_{0j}, p_{1j}, \ldots, p_{lj})\,dt
 \end{multline}
 
\noindent and where $v_j(t; t_{0j}, p_{1j}, \ldots, p_{lj})$ is the bell-shaped velocity function of stroke $j$, $t$ the time, $t_{0j}$ the time of stroke occurrence, $p_{lj}$ the parameter $l$ of the velocity function and $D_j$ the amplitude of the stroke.

\begin{figure*}
\centering
 \includegraphics[width=0.8\linewidth]{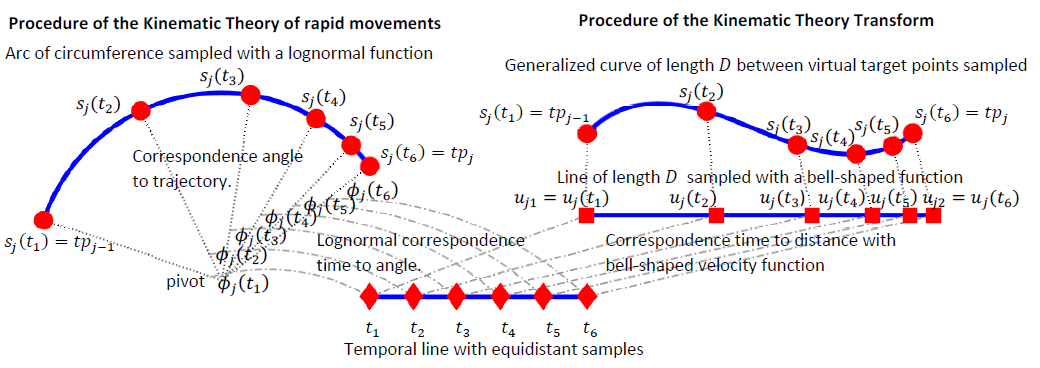}
  \caption{Procedure to sample the generalized link between virtual target points of a primitive. Example particularized for the case of Clothoid and Lognormal.}
  \label{fig4} 
\end{figure*}

The solution to this equation holds for $u_{j1}\leq u_j(t) \leq u_{j2}$. In this case, instead of using the Sigma-Lognormal model equations to reconstruct a trajectory (see Supplemental file), the iDeLog method uses Eqs.~\eqref{eq14}, ~\eqref{eq15} and ~\eqref{eq16}. Fig.~\ref{fig4} illustrates the underlying idea of this procedure.

The introduction of new trajectories between virtual target points which differ from arcs, implies a modification in the method that works out $\theta_{sj}$ and $\theta_{ej}$. Now, these two values are defined as follows:

\noindent     1. The angle $\theta_{sj}$ is obtained as the angle at $sp_{j-1}$ of the circumference that traverses the points $sp_{j-1}, mp_{j1}$  and $mp_j, mp_{j1}$. \textcolor{black}{Where $mp_j$ denotes the midpoints of the circular arcs, which are identified in the middle of the trajectory between $sp_{j-1}$ and $mp_j$.}


\noindent    2. The angle $\theta_{ej}$ is obtained as the angle at $sp_{j}$ of the circumference that traverses the points $mp_j, mp_{j2}$  and  $sp_{j}, mp_{j2}$,  being the point in the middle of the trajectory between $mp_j$ and $sp_j$.

 This new procedure offers the possibility of linking the virtual target points with any curve, when it is defined by its parametric equations. Obviously, its use can be also extended to arcs of circumferences. In the next section we particularize this general procedure for clothoid curves because they can be defined by virtual points and tangent angles and can include inflexion points. 

 \subsection{Clothoids as curves that link virtual target points}
 
 Clothoids are an example of curves that can link virtual target points in the KTT. They represent a useful option because of the parameters required for their definition: their starting and ending points, as well as their starting and ending angles, fit perfectly with the KTT.

 A clothoid is a curve whose shape changes linearly with its curve length; however, the shape of a circular arc is equal to the reciprocal of the radius. Thus, the transition from $\theta_{sj}$ and $\theta_{ej}$ can be smoother than in the case of a circumference and may include an inflection point. 
 
 In addition to its use for modelling handwriting graffiti~\cite{berio2017computer}, this kind of curve optimizes the acceleration and jerk of ballistic trajectories, which are a characteristic of biological trajectories~\cite{arechavaleta2008optimality}. Clothoids are also commonly referred to as Spiros, Euler spirals, or Cornu spirals. 

 Clothoids are defined by the following system of ordinary differential equations~\cite{bertolazzi2015g1} for each primitive: 


\begin{equation} 
x_p  = \displaystyle \int_0^u \cos(\pi v^2/2)\,dv; \qquad  y_p  = \displaystyle \int_0^u \sin(\pi v^2/2)\,dv
\end{equation}

The process of obtaining a clothoid, given two consecutive virtual target points $tp_{j-1}$ and $tp_{j}$ and the corresponding two angles $\theta_{sj}$ and $\theta_{ej}$, can be conducted using a software package\footnote{\url{https://es.mathworks.com/matlabcentral/fileexchange/42113-ebertolazzi-g1fitting}. Accessed 23 Aug. 2022.}.

\begin{figure}
\centering
 \includegraphics[width=0.7\linewidth]{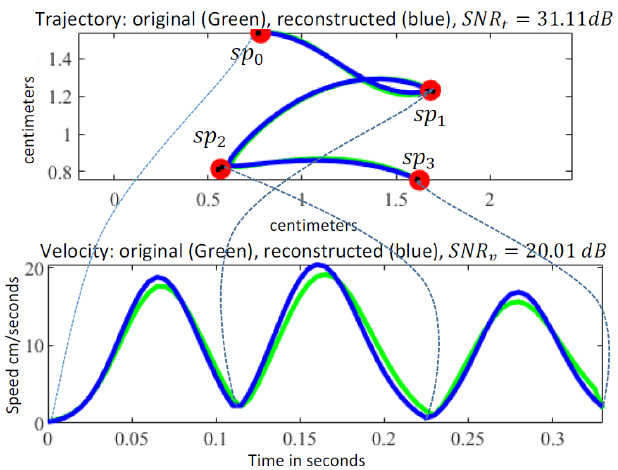}
  \caption{Same example as Fig.~\ref{fig3} but reconstructing the trajectory with clothoids }
  \label{fig5} 
\end{figure}

As an example, Fig.~\ref{fig5} illustrates a better reconstruction of the trajectory when clothoids are used when compared to Fig.~\ref{fig3}, where only arcs of circumferences were used. The reconstruction of the velocity profile is not as good as with circle arcs though.

 \section{Generalization of the velocity bell shaped curve}
 
 The musculoskeletal system is responsible for the movement of mammals, which is produced by the contraction of the skeletal muscles as a response to an action potential or electrical impulses from the nervous system. A skeletal muscle refers to multiple bundles of cells joined together called muscle fibers. The coordinated action of each muscle fiber produces the necessary strain to shorten the muscle and produce movement with a bell-shaped velocity profile.
 
 In motor control theory, various computational models have been developed to describe these velocity profiles~\cite{plamondon1993modelling}. Although it has been demonstrated that the Kinematic Theory and its Sigma-Lognormal model could be seen as the ultimate minimization model for human movement, several other analytical models might be of interest in some biological applications and in various fields. Most of these alternative models, by following the biological procedure of action potentials which excite muscular fibers, consider the muscular fiber as a subsystem of the skeleton muscle, which is considered as a system in itself. Thus, the movement is modeled as the impulse response to a set of $L$ combined subsystems to the sequence of motoneuron commands~\cite{plamondon1989evaluation}. If these subsystems are considered as independent, a symmetric bell-shaped velocity profile emerges. If the synergetic coupling of numerous neuromuscular subsystems is taken into account, then asymmetric lognormal profiles emerge~\cite{djioua2010limit}.
 
 Given the number of different proposed shapes for the velocity profile~\cite{plamondon1993modelling} and the similarity among them, we have implemented and tested the following bell-shaped functions.
 
 \subsection{Gaussian function}
 
 Whether all the subsystems are considered independent and the number of subsystem tends toward the infinite, based on the Central Limit Theorem, the bell-shaped velocity profile of a movement can be approximated by a symmetric Gaussian function defined as:
 
 \begin{equation}
  v_j(t;\mu_j,\sigma_j^2) = \dfrac{D_j}{\sigma_j\sqrt{2\pi}} \text{exp} \left\lbrace \dfrac{-(t-\mu_j)^2}{2\sigma_j^2}\right\rbrace
 \end{equation}

 \noindent where $D_j$ denoted the area under the velocity bell curve, $\mu_j$ the mean and $\sigma_j^2$ the variance. It could be expected that the mean of the Gaussian function would be around the peak of the velocity bell. This function is supported in $(-\infty, +\infty)$ and it is symmetric.
 
 \subsection{Gamma function}
 
 Although the Gaussian function is a good approximation of the bell-shaped velocity peaks for biological movements, it is well-known that the velocity peaks are asymmetric: they start at the time $t_{0j}$ and there are not infinite subsystems. As such, the minimum time theory model represents the velocity response of the neuromuscular system by the convolution product of a large number of first-order low-pass filters. In this case, the response tends to a special case of the Gamma function~\cite{djioua2010limit} which is asymmetric. Specifically, the Gamma function used here to fit the velocity bell is defined by:
 
 \begin{equation}
  v_j(t; t_{0j}, \alpha, \beta) = D_j \dfrac{(t-t_{0j})^{\alpha-1}\text{exp}\{-(t-t_{0j})/\beta\}}{\beta^\alpha \displaystyle \int_0^\infty t^{\alpha-1}\text{exp}(-t)\,dt}
 \end{equation}

 \noindent where $\alpha$ and $\beta$ are called the shape and scale parameter, respectively. This function is defined for $t_{0j}<t<\infty$ and $\alpha, \beta>0$. From the computational point of view, it adds a new parameter or degree of freedom for fitting the velocity bell curve. 
 
 \subsection{Beta function}
 
 The minimum jerk theory generalizes the expression for the velocity bell curve as a smooth Beta model, which is itself a particular case of the Gamma function~\cite{djioua2010limit}. In this case, the velocity bell shape is double bounded, i.e. it starts at $t_{0j}$ and ends at $t_{ej}$. The beta function is defined by:
 
 \begin{equation}
  v_j(t; t_{0j}, \alpha, \beta) = D_j \dfrac{(t-t_{0j})^{\alpha-1}(1-(t-t_{0j}))^{\beta-1}}{\displaystyle \int_0^1 u^{\alpha-1} (1-u)^{\beta-1}\,du}
 \end{equation}
 
 \noindent where $\alpha$ and $\beta$ are the first and second shape parameters. This function is defined for $t_{0j}<t<t_{0j}+1$, and $\alpha,\beta>0$. Because of the inherent finite time length of the beta velocity bell shape curve, $t_{min\,j}-t_{0j}<1$, is required to hold, which is not a practical solution if $t_{min\,j-1}-t_{0j}\approx0.5$~\cite{ferrer2018idelog}.

 \subsection{Lognormal function}
 
 In the real world, neuromuscular subsystems are physically connected and system dependency cannot be neglected. The Central Limit Theorem can be used to show that the velocity shape tends to a lognormal impulse response when the number of subsystems tends towards the infinite. The lognormal function span is $t_{0j}<t<\infty$.
 
 \subsection{Double-bounded lognormal}
 
 This function introduces an extension to the lognormal infinite length response to allow for both a lower and an upper bound to the values of the lognormal. This extension is called the four-parameter distribution in~\cite{10.2307/2227716}, or the double bounded lognormal in~\cite{plamondon1993modelling} and confines the lognormal to the range $t_{0j}<t<t_{ej}$ by assuming that the rate $(t-t_{0j})/(t_{ej}-t)$ is lognormal. As a result, the double bounded lognormal is defined as:
 
 \begin{equation}
  v_j(t; t_{0j}, t_{ej}, \mu_j,\sigma_j^2)=\dfrac{D_j(t_{ej}-t_{0j})}{\sigma_j\eta(t)\sqrt{2\pi}}\text{exp}\left\lbrace \dfrac{-[\text{ln}\zeta(t)-\mu_j]^2}{2\sigma_j^2}\right\rbrace
 \end{equation}
 
 \noindent where \textcolor{black}{$\zeta(t) = (t-t_{0j})/(t_{ej}-t)$ and $\eta(t) = (t-t_{0j})\cdot(t_{ej}-t)$. } The variables $D_j, t_{0j}, \mu_j$ and $\sigma_j^2$ are the same as for the lognormal and $t_{ej}$ is the end time of the lognormal. In this case, we have to estimate an additional parameter, i.e. $t_{ej}$. This is, therefore, an additional degree of freedom which can lead to a better adjustment of the velocity bell curve.

 \subsection{Generalized Extreme Value function (GEV)}
 
 Different velocity profile models are needed when the number of the subsystems (i.e. the number of muscle fibers involved in the human action) is large, specifically when they tend to the infinite. As a result, it could be said that the velocity bell shape tends to the lognormal, but there could be some deviations, depending on the finite number of muscle fibers involved in the movement and their correlation.
 
 In this context, the KTT method provides the possibility of fitting the velocity bell shape with a Generalized Extreme Value (GEV) function. GEV had been successfully applied to model physical and biological phenomena. This is a particular case of the Central Limit Theorem for sums of strongly correlated systems~\cite{bertin2006generalized}. GEV combines three simple distributions into a single form, giving a continuous range of possible shapes. As a consequence, the GEV leads to {\it ``let the data decide''} which distribution is most appropriate for each primitive. The GEV is defined as:
 
 \begin{equation}
  v_j(t; t_{0j},\xi_j,\mu_j,\sigma_j^2)=\dfrac{D_j}{\sigma_j}s(t-t_{0j})^{\xi_j+1} e^{-s(t-t_{0j})}
 \end{equation}

 \noindent where
 
 \begin{equation}
  s(t)= 
\begin{cases}
    \left[1+\xi_j \left(\dfrac{t-\mu_j}{\sigma_j} \right) \right]^{-1/\xi_j},& \text{if } \xi_j \neq 0\\
    \text{exp}[-(t-\mu_j)/\sigma_j],              & \text{if } \xi_j = 0
\end{cases}
 \end{equation}
 
 \noindent $mu_j$ is the location parameter, $\sigma_j$ the scale parameter and $\xi_j$ the shape parameter. This governs the tail behavior and identifies the bell shape as belonging to one of the three sub-families of distributions:
 

\noindent 1. If $\xi_j=0$, then the GEV is a Gumbel distribution, which is unbounded.
\noindent 2. If $\xi_j>0$, then the GEV is a Fréchet distribution, with its lower tail bounded and with a large upper tail.
\noindent 3. If $\xi_j<0$, then the GEV is a Weibull distribution, with its upper tail bounded and short.
 
 As can be seen, the GEV has the same number of parameters as the double bounded lognormal.
 
\begin{figure}[!ht]
\centering
 \includegraphics[width=.96\linewidth]{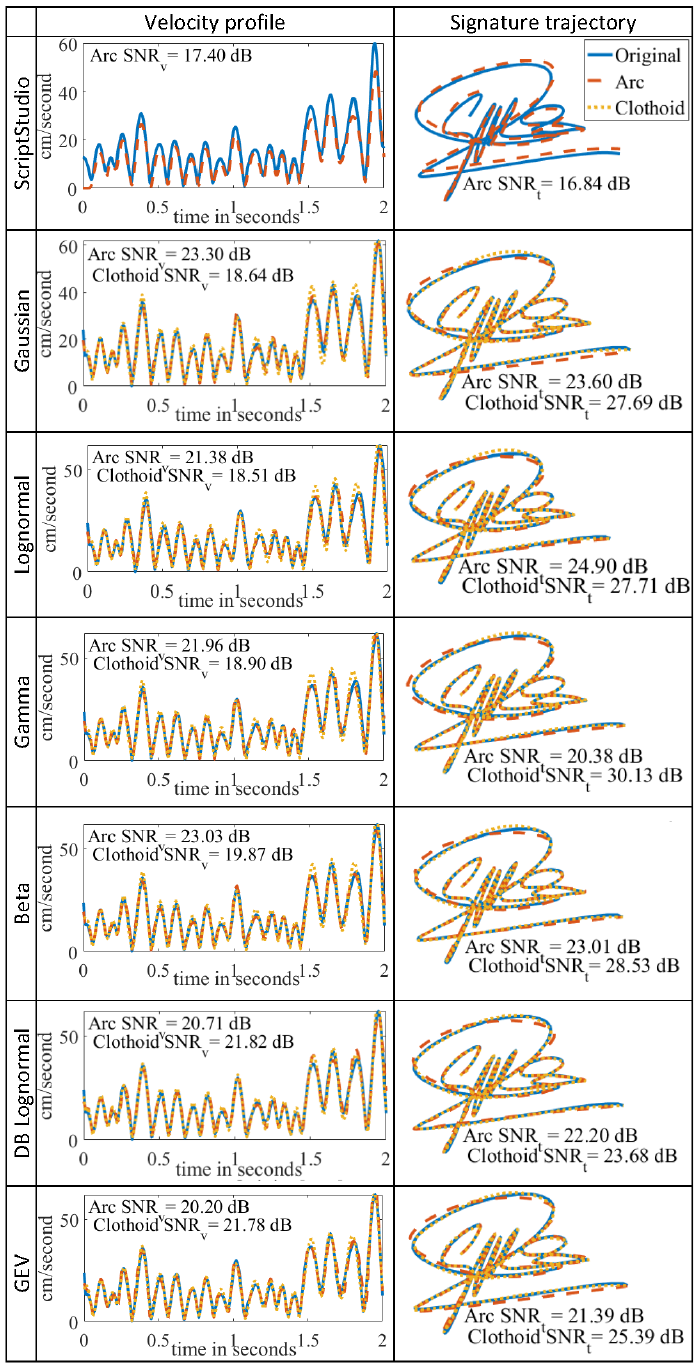}
  \caption{Comparison between the original velocity and the reconstructed velocity with Script Studio and iDeLog with the KTT.}
  \label{fig6} 
\end{figure}

 An example of velocity profiles reconstructed with each of these velocity bell models can be seen in Fig.~\ref{fig6}. Evidently, the clothoid reconstruction is better than the arc for the trajectory but worse for the velocity, except with double bounded lognormal and GEV, both of which have a supplementary degree of freedom.

 \section{KTT Evaluation In Complex Movements}
 
 
 
  In this section, we first introduce the database used for the experiments. Next, we analyze the reconstruction performances with arcs and clothoids  by using the different bell-shapes for velocity curve modelling to assess the strengths and weaknesses of each KTT configuration. Finally, confirmatory data analysis was performed for assessing the meaningful statistical relationships between the different approaches for reconstructing both the trajectory and velocity.

 \subsection{Database collection of different movements}


 We evaluated the KTT in human, animal and robotic movements.   All  data were acquired by entirely different sensors, such as tablet styli, interactive whiteboards and inertial systems.

%
 
 \begin{table*}[h!]
   \caption{\textcolor{black}{Comparison between different KTT configurations for all database in terms of Average (STD). }}
   \label{t1}
\centering
\scriptsize
 \begin{tabular}{|c|c|llll|llll|} 
 \hline
 \multirow{2}{*}{\tiny Database}&\multirow{2}{*}{Velocity Bell}&\multicolumn{4}{c|}{Using Circumferences}&\multicolumn{4}{c|}{Using Clothoids}\\ \cline{3-10}
 & &$SNR_t$(dB)&$SNR_t/N$&$SNR_v$(dB)&$SNR_v/N$&$SNR_t$(dB)&$SNR_t/N$&$SNR_v$(dB)&$SNR_v/N$\\ \hline\hline
 \multirow{6}{*}{\rotatebox{90}{BiosecureID}}&Gaussian&21.10(5.63)&0.82(0.65)&19.22(1.16)&0.73(0.37)&22.71(6.27)&0.88(0.73)&19.24(1.22)&0.73(0.38)\\
&Lognormal&21.10(5.67)&0.82(0.65)&18.94(1.19)&0.72(0.38)&22.64(6.25)&0.88(0.73)&18.92(1.23)&0.72(0.39)\\ 
&Gamma&21.15(5.68)&0.82(0.65)&19.07(1.19)&0.73(0.38)&22.69(6.20)&0.88(0.73)&19.07(1.21)&0.73(0.39)\\ 
&Beta&21.09(5.63)&0.81(0.65)&19.17(1.09)&0.73(0.39)&22.53(6.20)&0.87(0.73)&19.20(1.10)&0.73(0.40)\\ 
&DBL&20.49(5.66)&0.79(0.64)&18.80(1.37)&0.71(0.42)&22.14(6.26)&0.86(0.73)&18.90(1.40)&0.72(0.43)\\ 
&GEV&21.08(5.66)&0.81(0.65)&18.73(1.66)&0.71(0.44)&22.66(6.27)&0.87(0.73)&18.77(1.70)&0.71(0.46)\\ \hline
\multirow{6}{*}{\rotatebox{90}{MCYT}}&Gaussian&20.00(5.31)&0.79(0.93)&17.94(1.20)&0.69(0.57)&21.38(6.18)&0.84(1.03)&17.86(1.24)&0.69(0.59)\\ 
&Lognormal&19.96(5.30)&0.79(0.93)&17.74(1.21)&0.68(0.57)&21.36(6.22)&0.84(1.02)&17.66(1.27)&0.68(0.58)\\ 
&Gamma&19.88(5.30)&0.78(0.93)&17.82(1.25)&0.69(0.58)&21.37(6.20)&0.85(1.02)&17.72(1.28)&0.68(0.59)\\ 
&Beta&19.93(5.29)&0.79(0.93)&17.94(1.15)&0.69(0.60)&21.34(6.21)&0.84(1.03)&17.88(1.18)&0.69(0.61)\\ 
&DBL&19.70(5.36)&1.04(0.94)&17.78(1.46)&0.91(0.67)&21.10(6.18)&1.15(1.02)&17.81(1.47)&0.93(0.67)\\ 
&GEV&19.44(5.24)&0.77(0.93)&17.35(1.77)&0.66(0.68)&20.89(6.20)&0.83(1.02)&17.30(1.74)&0.66(0.70)\\ \hline
\multirow{6}{*}{\rotatebox{90}{Handwriting}}&Gaussian&23.02(6.76)&9.63(7.98)&17.68(5.83)&7.80(7.56)&23.82(7.09)&10.15(8.73)&17.45(5.93)&7.76(7.69)\\ 
&Lognormal&23.03(6.80)&9.65(8.06)&17.66(5.71)&7.81(7.48)&23.75(7.09)&10.14(8.76)&17.39(5.73)&7.74(7.51)\\ 
&Gamma&23.03(6.71)&9.63(7.92)&17.64(5.58)&7.79(7.34)&23.74(6.99)&10.11(8.62)&17.40(5.59)&7.72(7.38)\\ 
&Beta&23.11(7.58)&9.71(8.70)&17.76(7.01)&7.88(8.48)&23.89(7.25)&10.18(8.80)&17.48(6.05)&7.77(7.68)\\ 
&DBL&22.95(6.88)&9.64(8.11)&17.57(5.79)&7.73(7.48)&23.83(7.23)&10.17(8.83)&17.40(5.88)&7.70(7.60)\\ 
&GEV&22.85(7.38)&9.61(8.53)&17.22(6.32)&7.59(8.00)&23.49(7.31)&10.02(8.66)&16.94(5.60)&7.47(7.12)\\ \hline
\multirow{6}{*}{\rotatebox{90}{Robot}}&Gaussian&25.69(3.94)&3.07(1.16)&21.47(1.63)&2.55(0.82)&26.27(3.71)&3.15(1.18)&21.57(1.54)&2.56(0.81)\\ 
&Lognormal&26.98(3.61)&3.23(1.16)&21.50(1.58)&2.56(0.81)&26.65(3.72)&3.19(1.17)&21.54(1.69)&2.55(0.79)\\ 
&Gamma&26.42(3.73)&3.16(1.16)&21.60(1.76)&2.56(0.81)&26.27(3.88)&3.16(1.23)&21.65(1.57)&2.57(0.80)\\ 
&Beta&26.53(3.77)&3.16(1.12)&21.42(1.63)&2.54(0.80)&26.13(3.66)&3.14(1.16)&21.47(1.52)&2.55(0.79)\\ 
&DBL&25.83(4.39)&3.12(1.29)&20.76(1.71)&2.47(0.79)&26.24(4.33)&3.17(1.30)&20.92(1.54)&2.49(0.80)\\ 
&GEV&26.69(3.99)&3.21(1.20)&20.14(1.67)&2.39(0.77)&26.22(3.73)&3.13(1.11)&20.15(1.40)&2.39(0.75)\\ \hline
\multirow{6}{*}{\rotatebox{90}{Dogs}}&Gaussian&26.09(5.72)&1.44(0.45)&18.84(2.57)&1.02(0.20)&28.00(5.27)&1.55(0.45)&18.93(2.60)&1.03(0.20)\\ 
&Lognormal&26.26(5.64)&1.45(0.46)&18.97(2.38)&1.03(0.21)&28.18(5.37)&1.56(0.45)&19.03(2.47)&1.04(0.21)\\ 
&Gamma&26.36(5.65)&1.45(0.46)&19.02(2.42)&1.04(0.20)&28.08(5.40)&1.56(0.46)&19.15(2.54)&1.05(0.20)\\ 
&Beta&25.94(5.61)&1.43(0.44)&18.93(2.61)&1.03(0.19)&28.04(5.26)&1.55(0.43)&19.07(2.56)&1.04(0.19)\\ 
&DBL&25.46(5.47)&1.40(0.43)&18.46(2.27)&1.00(0.18)&27.54(5.30)&1.52(0.43)&18.63(2.26)&1.02(0.18)\\ 
&GEV&26.16(5.65)&1.44(0.46)&19.10(2.13)&1.04(0.21)&27.92(5.46)&1.55(0.46)&19.14(2.09)&1.05(0.21)\\ \hline 
\multicolumn{10}{l}{\scriptsize $^*N$ denotes the number of peaks in the bell-shaped velocity profiles.}\\
 \end{tabular}
\end{table*}

\begin{figure}
\centering
 \includegraphics[width=0.75\linewidth]{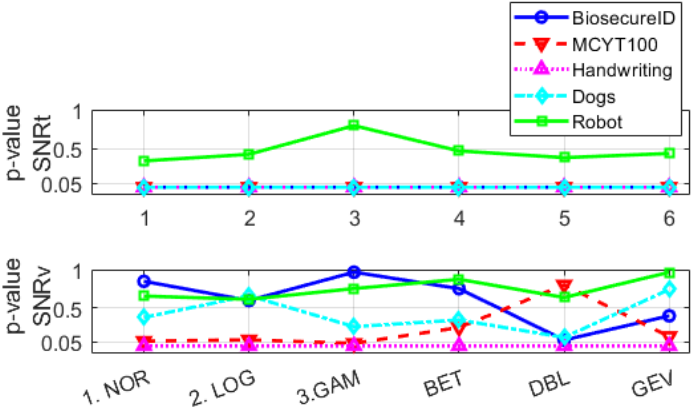}
  \caption{P-values results of the non-parametric Mann-Whitney U-test. Comparison in terms of $SNR_t$ and $SNR_v$ between arc of circumference and clothoid across all the databases and bell-shaped functions.}
  \label{fig8} 
\end{figure}

\begin{figure}
\centering
 \includegraphics[width=0.75\linewidth]{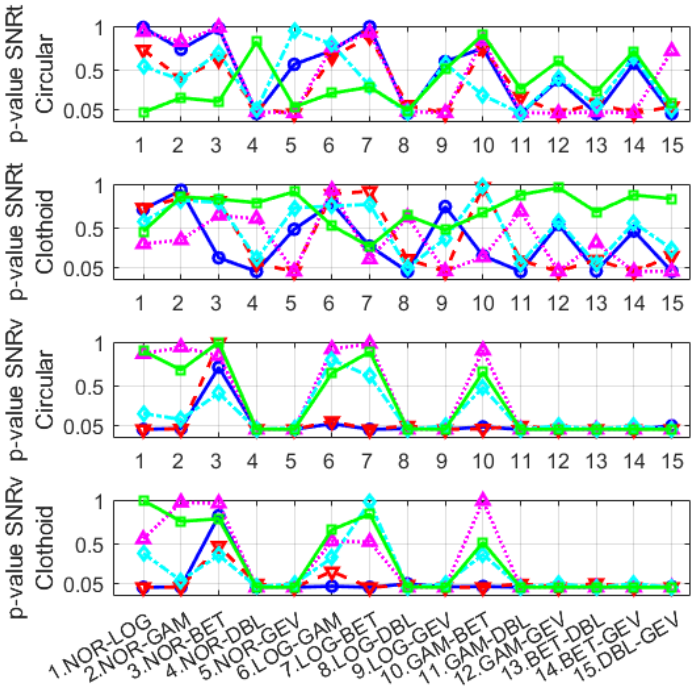}
  \caption{P-values results of the non-parametric Mann-Whitney U-test. $SNR_v$  and $SNR_t$ parameter comparison across all pair combination of bell-shaped functions and databases.}
  \label{fig9} 
\end{figure}

	In the case of human movements, we used two handwritten signatures databases (BiosecureID~\cite{galbally2015line} and  MCYT100~\cite{ortega2003mcyt}) and  handwriting on a whiteboard database (IAM On-Line Handwriting Database~\cite{liwicki2005handwriting}).
	
	For animals, eleven dogs, when walking without any constraints during more than 40 minutes of recording, were used. A harness, attached to the dog's back, was used to retain a wireless Neuron MOCAP sensor.

 Finally, we recorded the movement of an anthropomorphic robotic arm when writing. The robot wrote 100 doodles on a Wacom Intuos Pro A4 tablet with an attached WACOM ballpoint pen.

 \subsection{Experimental performance \textcolor{black}{and statistical study}}
 
 Both $SNR_v$ and $SNR_t$ were used to evaluate the capacity of the KTT implemented in iDeLog for reconstructing a particular movement. Better results in $SNR_v$ and $SNR_t$ indicate that the movement was better approximated by the bell-shaped function and trajectory curves used in the reconstruction. We also computed the ratio $SNR_v/N$ and $SNR_t/N$, $N$ being the number of peaks in the bell-shaped velocity functions. This ratio measures the efficiency of the KTT, that is to say, how many bell-shaped velocity functions are required to reach the obtained $SNR_v$ and $SNR_t$. In this case, the greater this ratio, the more efficient is the KTT representation~\cite{laniel2020kinematic}. We consider that the KTT provides a good approximation to the spatiotemporal sequence if the $SNR_v$ and $SNR_t$ are individually greater than 15dB~\cite{o2009development}.

Table~\ref{t1} shows the averaged $SNR_v$ and $SNR_t$ results with all the databases as well as the ratio of $SNR_v/N$  and $SNR_t/N$ \textcolor{black}{and their standard deviations}. It is worth pointing out that the results provided compare the reconstructed spatiotemporal sequence with the original one, i.e. the raw signal provided by the devices without smoothing them.

 Also, a statistical study was carried out to confirm these findings. Firstly, a Jarque-Bera test at 5\% significance was applied. The null hypothesis of this test assesses whether a distribution is normally distributed by working out the upper tail probability of the chi-squared distribution, defined by the skewness and kurtosis of the raw data. For $SNR_v$, the null hypotheses were rejected in all cases. For $SNR_t$, the hypothesis was rejected in the majority of the cases. Consequently, non-parametric tests were applied for a fair comparison. 
 
 As we compared pairs of sequences, the statistical similarity was assessed through the non-parametric Mann-Whitney U-test. The null hypothesis is that two different parameter distributions have no relationship between them. This can be rejected by observing that the p-value is greater than $p>0.05$. Also, we observe how strong the statistical relationship is in the pair of studied parameters.

	In the case of using clothoid curves instead of circumference arcs, excellent improvements can be observed in $SNR_t$ and $SNR_t/N$ while  $SNR_v$ and $SNR_v/N$ barely changes. \textcolor{black}{Thus, a slightly higher standard deviation can be found in $SNR_t$, as compared to $SNR_v$, for handwriting and signatures. However, the $SNR_t/N$  or $SNR_v/N$  ratios are very similar when arcs or clothoids are used. These findings are confirmed in the statistical analysis in Fig.~\ref{fig8}. We observe that changing from arcs of circumference to clothoids affects $SNR_t$, but barely changes $SNR_v$.}

 \textcolor{black}{Furthermore, we can compare how the $SNR_t$ and $SNR_v$ vary for each dataset. We got similar performances on all functions for signatures (MCYT100, and BiosecureID). However, slightly better averages were noted for velocity and trajectory reconstructions with the Gaussian and Beta functions. As can be seen in Fig.~\ref{fig9}, the functions are not significantly different. In all cases, the standard deviation was relatively stable across all functions. Moreover, we obtained $19.24(1.22)$dB and $17.94(1.20)$dB in the $SNR_v$ for MCYT100 and BiosecureID databases with the Gaussian function and using arcs and clothoids, respectively. These performances improve the previous results obtained with ScriptStudio and iDeLog~\cite{ferrer2018idelog} using lognormals.}

 \textcolor{black}{In the case of handwriting, the beta function reported slightly better SNR averages, with higher standard deviations for both the trajectory ($23.89(7.25)$dB) and velocity reconstructions ($17.48(6.05)$dB), using clothoids. Regarding the $SNR_v$, significant statistical differences were observed when comparing the Gaussian function to the Double-bounded lognormal or to the Generalized extreme value ($p<0.05$). Furthermore, Fig.~\ref{fig9} reveals that significant differences were also detected between the lognormal function and the Double-bounded lognormal or Generalized extreme value.}
 
\textcolor{black}{No clear preference regarding a given function was observed when analyzing robotic arm data in Table~\ref{t1}. The best result was seen with the Gamma function, and was very similar to that obtained with Lognormal, Gaussian and Beta functions. In all cases, the robotic data had a lower standard deviation compared to the other datasets. One of the main factors contributing to this was the consistent output generated by the robots. While a common relationship was found in the functions when $SNR_t$ was studied, almost all function combinations suggested no significant statistical relationship ($<0.05$).}

 \textcolor{black}{On dogs, the performances were very similar across the functions $(SNR_v\sim 19.00$dB, $SNR_t\sim28.00$dB), with the gamma and GEV functions achieving the best results. These latter functions reported statistical differences in the case of $SNR_t$, but no differences for $SNR_v$.  Once again, the standard deviation was similar in all cases for the trajectory and reconstructed velocity parameters. }


 \section{Conclusion \textcolor{black}{and Discussion}}

 \begin{table}
   \caption{\textcolor{black}{Relation between parameters and first and second moment in the used bell-shaped functions}}    
   \label{t_rel}
\centering
\scriptsize
 \begin{tabular}{|c|m{6cm}|} 
 \hline
 Name&Mean, variance and parameters \\ \hline\hline
\multirow{2}{*}{Gaussian}& $\mu_j = M_j$\\
& $\sigma^2_j = V_j$\\ \hline 
\multirow{4}{*}{Lognormal}& $M_j = \exp{(\mu_j+\sigma_j^2/2)}$\\
 &$V_j=[\exp(\sigma_j^2)-1]\exp{(2\mu_j+\sigma_j^2)}$\\ \cline{2-2}
 &$\mu_j=\log (M_j/\sqrt{V_j+M_j^2})$\\
 &$\sigma^2_j=\log (V_j/M_j^2+1)$\\\hline
\multirow{4}{*}{Gamma}&  $M_j=\alpha_j/\beta_j$\\
 &$V_j=\alpha_j/\beta^2_j$\\ \cline{2-2}
 &$\alpha_j=M_j^2/V_j$ \\
 &$\beta_j=M_j/V_j$ \\ \hline 
\multirow{4}{*}{Beta}&  $M_j= \alpha_j/(\alpha_j+\beta_j)$ \\ 
&  $V_j=\alpha_j\beta_j/[(\alpha_j+\beta_j)^2(\alpha_j+\beta_j+1)]$ \\ \cline{2-2}
&  $\alpha_j=[M_j(1-M_j)/V_j-1]M_j$\\
&  $\beta_j=[M_j(1-M_j)/V_j-1](1-M_j)$\\ \hline
\multirow{9}{1.4cm}{\centering Generalized extreme value}& $M_j = \begin{cases}
  \mu_j+\sigma_j(g_1-1)/\xi_j  &  \xi_j \neq 0, \xi_j<1 \\
  \mu_j+\sigma_j\gamma & \xi_j = 0
\end{cases}$ \\        
&$V_j = \begin{cases}
 \sigma_j^2(g_2-g_1^2)/\xi_j^2  & \xi_j \neq 0, \xi_j<0.5 \\
  \sigma_j^2\pi^2/6 & \xi_j = 0
\end{cases}$ \\
&$g_k = \Gamma(1-k\xi_j)$ and $\gamma$ is Euler's constant\\ \cline{2-2}
&$\sigma_j^2=\sqrt{6V_j}/\pi$\\
&$\mu_j=M_j-\sigma_j^2\gamma$\\
&Then $\xi_j$ is adjusted by hill-climbing starting $\xi_j=0$ and $\xi_j<0.5$\\ \hline 
\multirow{4}{1.4cm}{\centering Double-bounded lognormal}&$M_j\approx\exp{(\mu_j+\sigma_j^2/2)}$\\
&$V_j\approx[\exp{(\sigma_j^2)}-1]\exp{(2\mu_j+\sigma_j^2)}$\\ \cline{2-2} 
&Obtained by hill-climbing starting $\mu_j=0.6$, $\sigma_j^2=0.2$ and $t_e=t_{0_j}+(t_{\min,j}-t_{\min,j-1})$\\ \hline 
  \end{tabular}
\end{table} 

 We propose a generalization of the Kinematic Theory of Rapid Movement which we call the Kinematic Theory Transform (KTT). The KTT models the spatial and temporal information jointly in terms of trajectory and velocity. Beyond reproducing rapid movement as an overlapped combination of arc traversed at lognormal velocity, the KTT permits the use of any trajectory and bell-shaped functions to represent the velocity. As a proof of concept, we have studied the application of clothoids for the trajectory that links the virtual target points, along with six possible functions for the velocity: Gaussian, Lognormal, Gamma, Beta, Double Bounded Lognormal and GEV.

\textcolor{black}{This paper raises two points. The first is the biological justification and utility of the new primitives, while the second one concerns the biological meaning of the parameters of the new primitives.}

\textcolor{black}{Regarding the justification of the new primitives, we agree with the kinematic theory that has been proposed to study and analyze rapid human movements. We are aware that it is based on the central limit theorem, which predicts a lognormal impulse as a response to the behavior of a large number of interdependent neuromuscular networks. However, realistic human movements come from a limited number of joints, and not all of them are perfectly learned and synchronized. This concurs with our observation of imperfect lognormals in the databases analyzed. The new primitives constitute a tool for analytically analyzing the deviations due to these non-perfect fits of the conditions of the central limit theorem.}

\textcolor{black}{Regarding the biological meaning of the parameters, $SNR_v$, $SNR_t$, Number of Lognormals ($NbLog$), $SNR_v/Nblog$, and $SNR_t/NbLog$ explain the quality of the neuromotor control. These parameters have the same meaning across all functions and help to study the deviation of the movement carried out from a rapid and well-learned movement. Regarding the Neuromotor action plan parameters, $t_{0_j}$ and $D_j$ have similar numerical values in all proposed functions, and therefore maintain the same biological meaning. The parameters of the new bell-shaped function, such as $\alpha_j$ and $\beta_j$ for the gamma and beta functions, or $\zeta_j$, $\eta_j$, and $\sigma^2_j$, which are related to the lognormal $\mu_j$ and $\sigma^2_j$ through the first and second moments~\cite{bowman2004estimation} of the functions. These parameters explains the Motor program execution. These relationships, shown in Table~\ref{t_rel}, establish the biological meaning of the new parameters. Additionally, we have been careful to maintain the same parameters in almost all proposed functions (See eq. 6-9).}


 Our experiments improve the state-of-the-art performance in terms of $SNR_v$ and $SNR_t$ when several biological movements such as on-line signatures, handwriting on an interactive whiteboard, the movement of dogs along with the movement of robotic arms, are reconstructed with the KTT. Confirmatory data analysis was carried out to assess the statistical differences between the solutions generated by multiple primitives. The KTT mathematical framework has been integrated into an improved version of iDeLog~\cite{ferrer2018idelog}. The extended iDeLog is freely distributed as a Matlab toolbox under a non-commercial research license agreement. This is a further advance in techniques for modelling spatiotemporal sequences when the velocity curve is bell-shaped. In addition to enriching the iDeLog method with new trajectories and bell-shaped functions in the present work, future efforts will be oriented to extend this framework to reconstruct 3D movements\textcolor{black}{~\cite{fischer2021modeling, schindler2018extending}.}

\section*{Acknowledgment}

This research is   supported by the Spanish MINECO (PID2019-109099RB-C41 project), the European Union (FEDER program), and the CajaCanaria and la Caixa bank grant 2019SP19.  Authors would like to thank Dra. Elena Carreton from Veterinary Clinical Hospital at Universidad de Las Palmas de Gran Canaria for her assistance during the data collection from the dogs.
 

%
%
%
%

\bibliography{main}

\clearpage

%
%
%
%
%
%

\end{document}